\documentclass{article}

 \usepackage[preprint]{neurips_2020}

\usepackage[utf8]{inputenc} 
\usepackage[T1]{fontenc}    
\usepackage{hyperref}       
\usepackage{url}            
\usepackage{booktabs}       
\usepackage{amsfonts}       
\usepackage{nicefrac}       
\usepackage{microtype}      
\setcitestyle{square,aysep={,},citesep={;}}
\usepackage{multirow}
\usepackage[pdftex]{graphicx}  

\title{Enhance Long Text Understanding via Distilled Gist Detector from Abstractive Summarization}

\author{Yan Liu, Yazheng Yang \\
  College of Computer Science, Zhejiang University \\
  {\tt $\{$yanliu1205, yazheng\_yang$\}$}}

\begin{document}

\maketitle

\begin{abstract}
  Long text understanding is important yet challenging in natural language processing. A long article or essay usually contains many redundant words that are not pertinent to its gist and sometimes can be regarded as noise. In this paper, we consider the problem of how to disentangle the gist-relevant and irrelevant information for long text understanding. With distillation mechanism, we transfer the knowledge about how to focus the salient parts from the abstractive summarization model and further integrate the distilled model, named \emph{Gist Detector}, into existing models as a supplementary component to augment the long text understanding. Experiments on document classification, distantly supervised open-domain question answering (DS-QA) and non-parallel text style transfer show that our method can significantly improve the performance of the baseline models, and achieves state-of-the-art overall results for document classification.
\end{abstract}

\section{Introduction}

Neural approaches have attracted significant popularity in the field of NLP over the past few years. 
Long text understanding has become an integral part of many NLP tasks such as document classification \citep{yang2016hierarchical,liu2017adversarial,li2018deep}, text summarization \citep{zeng2016efficient,nallapati2017summarunner}, and machine reading comprehension (MC) \citep{xiong2016dynamic,seo2016bidirectional,cui2017attention}, \emph{etc}.
For a long text, it intrinsically contains noise which is negligible to the main content of the text. Hence, it is vital to get a correct understanding that neural models focus on salient parts while neglect gist-irrelevant ones. 
Despite the promising results achieved by prior approaches, they typically train the neural model without disentangling the salient parts and negligible ones explicitly.

Abstractive text summarization, which is one of the significant tasks in natural language processing, aims to compress and rewrite a source text into a short version while retaining its main information. Recent approaches\citep{chopra2016abstractive,zeng2016efficient,nallapati2017summarunner} based on sequence-to-sequence (seq2seq) framework \citep{sutskever2014sequence} augmented with attention mechanism \citep{bahdanau2014neural} used for text summarization have achieved promising results. 
The hybrid pointer-generator network \citep{see2017get} takes the state-of-the-art of summarization task into a new level by taking advantage of both copying tokens from the source article and generating tokens from vocabulary distributions.

\begin{figure*}[t]
	\begin{center}
		\includegraphics[width=0.8\linewidth]{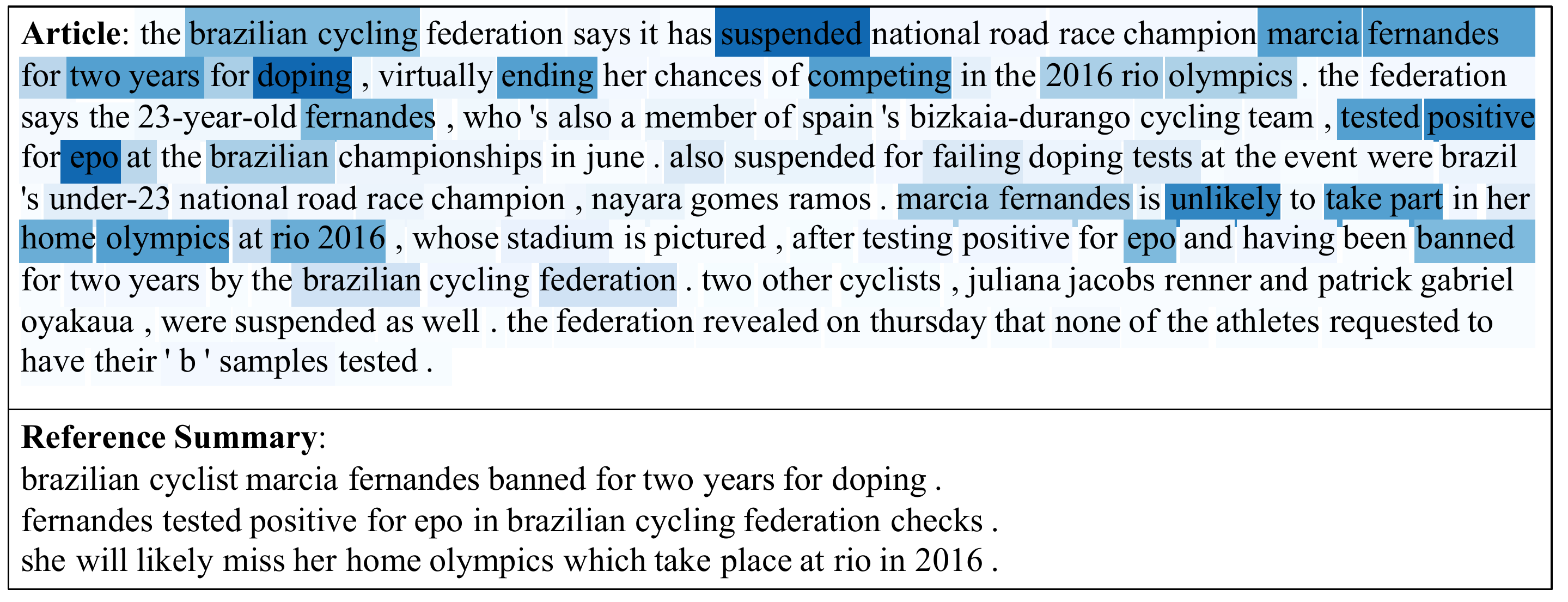}
		\caption{\label{fig1} An example from the CNN/Daily Mail dataset. The blue shading intensity represents the value of the average attention weight over all decoding steps from the abstractive summarization model.}
	\end{center}
\end{figure*}

The abstractive summarization model has the potential to find out the salient information for a long text.
Inspired by the effective information compression and the promising achievement of the recent approaches for the abstractive text summarization, we explore how to leverage the ability of gist detection (or salient information detection) from a summarization model and integrate it into existing models to enhance the long text understanding. 
Generating summarization for text requires to encode its words into a sequence of hidden representations firstly, then typical methods \citep{chopra2016abstractive,ranzato2016sequence,see2017get,guo2018soft} employ RNN-based decoder with attention mechanism to generate the target sequence word-by-word. On each decoding step, the decoder calculates the attention distribution and further gets a context vector, then produces the vocabulary distribution.
Here, the attention mechanism helps the abstractive summarization model focus on specific parts while decoding. Figure \ref{fig1} shows a simple example from the \emph{CNN/Daily Mail} \citep{nallapati2016abstractive} dataset, and the blue shading intensity represents the value of the average attention distribution. We observe that the abstractive summarization model learns to focus on gist-relevant parts while neglecting irrelevant ones.
Intuitively, the ability of gist detection can improve long text understanding through making existing models aware of the salient parts for a long text.
To utilize the gist detection ability, we use the average attention distributions over all decoding steps produced by the abstractive text summarization model as "soft targets" to train the student model with distillation mechanism. Rather than compressing models \citep{bucilu2006model,hinton2015distilling} to employ in same task, we match the "soft targets" directly with a different yet small network architecture and apply it to other NLP tasks.

To demonstrate the effectiveness of the proposed method, we verify it on three tasks:

(1) \emph{Document Classification}. We integrate the distilled gist detector into Document Classification model, which bases on the long shot-term memory network (LSTM) \citep{hochreiter1997long} and conduct experiments on 16 datasets of \cite{liu2017adversarial}. Our method achieves overall state-of-the-art results.

(2) \emph{DS-QA}. We apply the gist detector distilled from abstractive summarization model as a supplementary component to OpenQA model recently proposed by \cite{lin2018denoising}, which uses a paragraph selector and paragraph reader to predict the answer span. We evaluate the model augmented with gist detector on the \emph{SearchQA} \citep{dunn2017searchqa} and \emph{TriviaQA} \citep{joshi2017triviaqa} datasets and achieve significantly improvement of the performance.

(3) \emph{Non-parallel Text Style Transfer}. We modify the ARAE model proposed by \cite{kim2017adversarially}. 
To verify the effectiveness of our method, different from prior works that only focus on short sentences, we conduct experiments on longer texts collected from Yelp restaurant reviews and Amazon product reviews. 
The results show that our method obtains statistically significant improvement over the baseline model both on the sentiment accuracy and the content preservation.

\section{Related work}

\subsection{Long text understanding}

Long text understanding forms the foundation of many NLP tasks, range from document classification to machine reading comprehension. In general, these models take the pre-ordering process of mapping the input sequence of words in to a sequence of hidden states typically by RNN. The widely used document classifiers \citep{yang2016hierarchical,li2018deep,wang2018disconnected} take the end states or the summation of hidden states using attention as the holistic representation and then predict its label using a multi-layer perceptron (MLP). Further more, the initial decoder state for most RNN-based encoder-decoder models \citep{sutskever2014sequence,bahdanau2014neural,see2017get} is computed with the holistic representation. Similarly, after getting better understanding of the passage, many machine reading comprehension models \citep{xiong2016dynamic,seo2016bidirectional,cui2017attention} further employ another strategies to capture the interaction among the question and passage, and finally predict the answer span via pointer networks \citep{vinyals2015pointer} or MLP classifier.
We observe that people prefer to ask about the salient information of the passage. Intuitively, the gist detector could improve the MC model through focusing the salient parts of the passage. 

\subsection{Model compression \& transfer learning}

Model compression \citep{bucilu2006model,hinton2015distilling} transfers knowledge from cumbersome models to a smaller one applying in same task that is suitable for deployment while remaining competitive performance. The small model usually share same structure but less parameters.
Transfer learning, another way of utilizing existing knowledges, has been widely used among different tasks \citep{turian2010word,mccann2017learned,peters2018deep,devlin2018bert} in NLP.
The pre-trained word embeddings \citep{turian2010word} has become an essential part of NLP models. 
\cite{devlin2018bert} recently take the strategy of conditioning on both bidirectional context in all layers to pretrain BERT which is cumbersome, and achieve promising results after transferring to other NLP tasks by simply fine-tuning.
Different from prior works, we use different yet small model structure and transfer the distilled model to other NLP tasks for improving the performance of the baseline models.

\section{Methodology}

In this section, we first introduce the abstractive summarization model that is used as teacher model in this work. 
We then give an implementation of the gist detector which aims to learn the knowledge from the cumbersome teacher model. 
We further show the process of distillation and illustrate the strategy that how to integrate the gist detector into existing models. 

\subsection{Text summarization model}

The common abstractive summarization models adopt the encoder-decoder architecture in conjunction with attention mechanism. In such architecture, an encoder reads and maps the source text $x=\{ x_1,x_2,...,x_m \}$ into a sequence of vectors $ \mathbf{h}=\{ \mathbf{h}_1,\mathbf{h}_2,...,\mathbf{h}_m \}$, then the decoder models the probability of the target $y=\{ y_1,y_2,...,y_n \}$ conditioned on $\mathbf{h}$, where m and n are the length of x and y respectively.

\paragraph{Encoder} We first apply the word embedding along with character embedding via convolutional neural networks (CNN) to map the words of source article into a sequence of continuous representations. The character embedding has been proved to be useful in dealing with out-of-vocabulary (OOV) words \citep{yang2017words}. 
For encoding, we further feed the sequence of vectors into a two-layer bi-directional LSTM:
\begin{equation}
\mathbf{h}_i = {f_{enc}} (f_{wr}(x_i),\mathbf{h}_{i-1}), i= 1,...,m
\end{equation}
where $f_{wr}(w) = [{\rm Glove}(w)); {\rm Char}(w)]$ produces the word representation for each word $w$ in source article. Here, $ {\rm Glove}(w) $ is the word vector for $w$ from the lookup table of the pre-trained Glove word embeddings, $ {\rm Char}(w) $ is the character feature of $w$ produced by CNN,  $[;]$ represents concatenation, and $ {f_{enc}}(.) $ denotes the two-layer BiLSTM.

\paragraph{Decoder} 
On each decoding time step $t$, the decoder is fed with the word embedding of the previous word $y_{t-1}$. Here, we still use the pre-trained Glove.
We then pass $ \mathbf{e}_{t-1} $ into two-layer unidirectional LSTM and obtains the decoder state $\mathbf{s}_t$ for current time step. We calculate the attention distribution over the source sequence as in \citep{bahdanau2014neural}:
\begin{equation}
	\mathbf{e}_{t-1} = {\rm Glove}(y_{t-1})
\end{equation}
\begin{equation}
	\mathbf{s}_t = f_{dec}(\mathbf{e}_{t-1}, \mathbf{s}_{t-1})
\end{equation}
\begin{equation}
	d_{it} = \mathbf{v}_{a} {\rm tanh} (\mathbf{W}_{h} \mathbf{h}_i + \mathbf{W}_{h} \mathbf{s}_{t} + \mathbf{b}_a)
	\label{equation_summ_attn_logits}
\end{equation}
\begin{equation}
	\mathbf{c}_t = \sum_i \frac{{\rm exp}(d_{it})}{\sum_{j}{\rm exp}(d_{jt})} \mathbf{h}_i
\end{equation}
where $ \mathbf{v}_{a} $, $ \mathbf{W}_{h} $, $ \mathbf{W}_{h} $, and $ \mathbf{b}_a $ are all learnable parameters. Here, $\mathbf{c}_t$ is the context vector of the source sequence, which represents what has been read from the source-site on step $t$.
For the common abstractive summarization models, the context vector $\mathbf{c}_t$ is concatenated with decoder state $\mathbf{s}_t$, then projects to the vocabulary and produces the vocabulary distribution as follow:
\begin{equation}
	\mathbf{P}_v = {\rm softmax}(f_{map}([\mathbf{s}_t ; \mathbf{c}_t])
\end{equation}

To integrate the seq2seq model with copy mechanism, we use the soft switch gate proposed by \cite{see2017get} to combine the vocabulary distribution and copy distribution as formulation below:
\begin{equation}
	p_{gen} = {\rm \sigma}(\mathbf{W}_g[\mathbf{s}_t ; \mathbf{c}_t; \mathbf{e}_{t-1}] + \mathbf{b}_g)
\end{equation}
\begin{equation}
	P(w) = p_{gen} \mathbf{P}_v(w) + (1 - p_{gen}) P_c(w)
\end{equation}
here, $\mathbf{W}_g$ and $\mathbf{b}_g$ are trainable parameters, $\sigma$ is the sigmoid activator. $\mathbf{P}_v(w)$ is the probability of word $w$ from vocabulary distribution, $P_c(w)$ is the accumulation of attention weights if $w$ appears in the article words, otherwise it is 0.

\subsection{Gist detector}

There are many possible network architectures for the gist detector. 
A more abstractive view of the gist detector, that frees from any particular structure, is that it learns the knowledge from the teacher models about how to discriminate and predict the importance for each word in the source sequence.
The aim of this paper is not to explore this whole space but to show that a simple implementation for the gist detector works well and the distilled knowledge from abstractive summarization task augments the long text understanding for other NLP tasks.

Similar to previous section, we obtains the word representation via word embedding and character embedding, then pass it through a BiLSTM. 
To further relate the information of different positions in the sequence, we employ self-attention mechanism followed by residual connection:
\begin{equation}
	\mathbf{u}_i = {\rm BiLSTM} (f_{wr}(x_i),\mathbf{u}_{i-1})
\end{equation}
\begin{equation}
	\tilde{\mathbf{U}} = f_{sa} (\mathbf{U}) + \mathbf{U}
\end{equation}
Where $\mathbf{U}=\{ \mathbf{u}_1,\mathbf{u}_2,...,\mathbf{u}_m \}$. In our implementation, $f_{sa}$ is a multi-head scaled dot-product self-attention  that jointly attends to information from different subspaces of representation at different positions.
We then apply two-layer linear layers with a ReLU activation in between to produce the logits. The gist detector produces the probability (or strength of importance) for each word in the source article by further using a softmax output layer:
\begin{equation}
	\tilde{\mathbf{d}} = {\rm softmax}( \mathbf{V}^{'} {\rm ReLU}(\mathbf{V} \tilde{\mathbf{U}} + \mathbf{b}) + \mathbf{b}^{'} )
	\label{gist_output}
\end{equation}
where $\mathbf{V}$, $\mathbf{b}$, $\mathbf{V}^{'}$ and $\mathbf{b}^{'}$ are learnable parameters.

\subsection{Distillation process}

In the common form of distillation, the knowledge is transferred to the student model via matching a soft target distribution produced by the cumbersome model for each sample on the transfer set. 
When the correct labels are known for the transfer set, this method can further combine with the hard target that is the cross entropy with the correct labels.
In this work, since we only transfer the knowledge about how to find out the salient information, the distilled gist detector directly match the soft target.
\paragraph{The soft target}
We use the softmax layer with a temperature to convert the logits (formulated in equation (\ref{equation_summ_attn_logits})) produced by the summarization model into the attention distribution over the source article. To transfer this knowledge from the cumbersome model, the geometric mean of the attention distributions over all decoding steps is use as the soft target denoted as $Q$:
\begin{equation}
	Q(x_i) = \frac{1}{T_y} \sum_t \frac{{\rm exp}(d_{it} / T)}{\sum_{j}{\rm exp}(d_{jt} / T)}
	\label{equation_distill_target}
\end{equation}
where $T_y$ is the total decode steps, $T$ is the temperature. 

\paragraph{Training}
During Training, the gist detector use the same high temperature in softmax and obtains the predicted probability distribution $P_d$. Denote all the learnable parameters in the gist detector as $\Theta$, the transfer set as $D$. The objective of the training process is to learn the optimal parameters $\Theta^{'}$ that predicts the output distribution as similar as that from the teacher model. The cross entropy between two distributions servers as the distillation loss. The loss and objective can be formulated as follow:
\begin{equation}
	L_{KD}(D, \Theta) = - \sum_{D} \sum_{i=1}^{T_y} Q(x_i)log(P_d(x_i))
\end{equation}
\begin{equation}
	\Theta^{'} = \mathop{\arg\min}\limits_{\Theta} L_{KD}(D, \Theta)
\end{equation}

\subsection{Transfer strategy}

In most but a few cases, the dominant approaches for NLP problems project the text into a sequence of vectors that serves as the inner semantic representation for each word. These tasks shares similar pre-order processes where the neural networks first convert each word into vectors via word embeddings, then pass it through RNNs for obtaining high-level semantic information, and further compute the context representation by taking the end state or the summation of hidden states with attention mechanism. 
\begin{equation}
	\{ \mathbf{q}_1,...,\mathbf{q}_{T_x} \} = f_{project} ( \{ x_1,...,x_{T_x} \} )
\end{equation}
\begin{equation}
	\mathbf{v}_c = f_{c} ( \{ \mathbf{q}_1,...,\mathbf{q}_{T_x} \} )
	\label{equation_ctx}
\end{equation}
Given the context representation, the follow-up processes use task-specified methods, e.g. the classifier projects it into the class distributions via MLP, the encoder-decoder based model converts it as initial decoder state.
In such scenarios, we additionally produce gist-aware vector with the importance weights $\tilde{\mathbf{d}}=\{ \tilde{d}_1,...,\tilde{d}_{T_x} \}$ predicted by the gist detector in the equation (\ref{gist_output}) , and combine the two representation as follow:
\begin{equation}
	\mathbf{v}_{c}^{'} = (1 - \lambda) \mathbf{v}_{c} + \lambda \sum_i \tilde{d_i}\mathbf{q}_i
	\label{integrate_type_1}
\end{equation}
where $\mathbf{v}_{c}$ is the original context representation, $\lambda$ is a tunable parameter that adjusts the relative importance among the two context representations. 

In our work, providing the importance weights rather the context representation makes it possible that the gist detector uses less parameters and alleviates the impact of domain-specific information. The gist-aware vector is computed as the weighted summation of $\{ \mathbf{q}_1,...,\mathbf{q}_{T_x} \}$ which contains rich domain-specific information.
For extraction-like problems that predict scores for each position in source sequence, e.g. extractive QA, we integrate our gist detector in another way:
\begin{equation}
	r_{i}^{'} = (1 - \lambda^{'}) r_{i} + \lambda^{'} \tilde{d_i}
	\label{integrate_type_2}
\end{equation}
here, $ \lambda^{'} $ is tunable parameter, $r_{i}$ is the prediction of score for position $i$, e.g. the score produced by pointer networks in extractive QA.
\section{Experiments}

\subsection{Distillation}

To train the teacher models, we use the \emph{CNN/Daily Mail} \citep{nallapati2016abstractive} dataset. We follow the same setup and use the scripts\footnote{https://github.com/abisee/cnn-dailymail} provided by \cite{see2017get} to pre-process the data. We use the 100 one dimensional filters with width of 5 for CNN to capture the character features. We select the 300d Glove\footnote{https://nlp.stanford.edu/projects/glove/} pre-trained word embeddings and share the same word embedding weights between encoder and decoder. The hidden size of BiLSTM is 256. We take the Adam for optimization \citep{kingma2014adam} with learning rate (lr) of 0.0004, $\beta_1$ = 0.9, $\beta_2$ = 0.999. The dropout rate is set as 0.35, and batch size is 16. To avoid the gradient explosion problem, we apply the gradient norm clipping with maximum gradient norm of 2.0. We train an ensemble of 8 abstractive summarization models with average ROUGE ${\rm F}_{1}$ scores \citep{lin2004rouge} of 38.6, 16.3 and 35.4 for ROUGE-1, ROUGE-2 and ROUGE-L respectively. To construct the transfer set, the soft target is computed as the geometric mean of the attention distributions in the equation (\ref{equation_distill_target}), and the temperature is set as 4. For gist detector, we use 100d Glove for word embedding, 50d for character embedding, the hidden size for BiLSTM is 128. We take the same optimizing settings as that in teacher model.

\subsection{Application to document classification}

We first verify the effectiveness of our method on the document classification task. 
The BiLSTM have been a common approach for document classification task with competitive performance. 
The BiLSTM model concatenates the final state values of forward and backward pass as the context representation vector, then feed it into a MLP to predict the label. 
We take the BiLSTM as our baseline model and augment it with the gist detector as equation (\ref{integrate_type_1}). We use the accuracy as our evaluation metric. The datasets for our evaluation are the FDU-MTL\footnote{http://nlp.fudan.edu.cn/data/} datasets \citep{liu2017adversarial}.

\textbf{Implementation details:} We initialize the word embeddings with the 300d Glove.
The hidden size for BiLSTM is set as 256. The layer number of BiLSTM and MLP are both set as 2. We take the Adam as optimizer with lr = 0.001, $\beta_1$ = 0.9, $\beta_2$ = 0.999, 0.35 dropout and train for 6 epoches.
The $\lambda$ in equation (\ref{integrate_type_1}) is set to be 0.5 while integrating the BiLSTM model with our gist detector.
\begin{table}[tb]
	\centering
	\caption{Accuracies of our method, BiLSTM+GD on 16 datasets against competitive models. BiLSTM+NP uses the structure of gist detector but with random initialized parameters. The results of ASP-MTL \citep{liu2017adversarial}, S-LSTM \citep{zhang2018sentence}, Meta-MTL \citep{chen2018meta} are taken from corresponding paper, the results of S-LSTM are shown with rounding. The number in brackets represents the improvements relative to BiLSTM.}
	\begin{tabular}{p{1cm} c c c c c c}
		\hline 
		\multicolumn{1}{l}{\textbf{Datasets}}
		&\multicolumn{1}{c}{\textbf{ASP-MTL}}
		&\multicolumn{1}{c}{\textbf{S-LSTM}}
		&\multicolumn{1}{c}{\textbf{Meta-MTL}}
		&\multicolumn{1}{c}{\textbf{BiLSTM}}
		&\multicolumn{1}{c}{\textbf{BiLSTM+NP}}
		&\multicolumn{1}{c}{\textbf{BiLSTM+GD}}	\\
		\hline
		Appeal		&	87.0	 &	85.8 &	87.0 &	84.8 &	86.8\scriptsize{(+2.0)} &	\textbf{87.6}\scriptsize{(+2.8)}		\\
		Baby		&	88.2	 &	86.3 &	88.0 &	84.5 &	86.4\scriptsize{(+1.9)} &	\textbf{88.5}\scriptsize{(+4.0)}		\\
		Books		&	84.0	 &	83.4 &	\textbf{87.5} &	78.8 &	81.7\scriptsize{(+2.9)} &	86.7\scriptsize{(+7.9)}		\\
		Camera		&	89.2 &	90.0 &	89.7 &	86.3 &	89.1\scriptsize{(+2.8)} &	\textbf{90.8	}\scriptsize{(+4.5)}	\\
		DVD			&	85.5 &	85.5 &	\textbf{88.0} &	80.7 &	82.9\scriptsize{(+2.2)} &	87.8\scriptsize{(+7.1)}		\\
		Electronics	&	86.8 &	83.3 &	89.5 &	81.9 &	82.3\scriptsize{(+0.4)} &	\textbf{89.6}\scriptsize{(+7.7)}		\\
		Health		&	88.2 &	86.5 &	\textbf{90.3} &	83.0 &	84.5\scriptsize{(+1.5)} &	88.2\scriptsize{(+5.2)}		\\
		IMDB		&	85.5 &	87.2 &	88.0 &	79.8 &	81.2\scriptsize{(+1.4)} &	\textbf{88.1	}\scriptsize{(+8.3)}	\\
		Kitchen		&	86.2 &	84.5 &	\textbf{91.3} &	82.1 &	85.0\scriptsize{(+2.9)} &	90.7\scriptsize{(+8.6)}		\\
		Magazines	&	92.2 &	93.8 &	91.0 &	90.5 &	91.7\scriptsize{(+1.2)} &	\textbf{94.6	}\scriptsize{(+4.1)}	\\
		MR			&	76.7 &	76.2 &	77.0 &	76.1 &	76.5\scriptsize{(+0.4)} &	\textbf{78.2}\scriptsize{(+2.1)}		\\
		Music		&	82.5 &	82.0 &	86.3 &	79.6 &	82.9\scriptsize{(+3.3)} &	\textbf{86.4}\scriptsize{(+6.8)}		\\
		Software	&	87.2 &	87.8 &	88.5 &	85.4 &	87.8\scriptsize{(+2.4)} &	\textbf{90.3}\scriptsize{(+4.9)}		\\
		Sports		&	85.7 &	85.8 &	86.7 &	80.3 &	84.9\scriptsize{(+4.6)} &	\textbf{87.1}\scriptsize{(+6.8)}		\\
		Toys		&	88.0 &	85.3 &	\textbf{88.5} &	83.9 &	85.4\scriptsize{(+1.5)} &	88.3\scriptsize{(+4.4)}		\\
		Video		&	84.5 &	86.8 &	88.3 &	81.1 &	84.6\scriptsize{(+3.5)} &	\textbf{88.5}\scriptsize{(+7.4)}		\\
		\hline
		Overall		&	86.1 &	85.6 &	87.9 &	82.4 &	84.6\scriptsize{(+2.2)} &	\textbf{88.2}\scriptsize{(+5.8)}		\\
		\hline
	\end{tabular}
	\label{table_doc_classify}  
\end{table}

\textbf{Results:} As shown in the Table \ref{table_doc_classify}, the baseline model augmented with our gist detector obtains significant performance improvement on all of the datasets and outperforms prior approaches with overall accuracy of 88.2.
We further add the architecture of the gist detector to the BiLSTM model but initialize it from scratch. We observe that the overall performance drops 3.6, which indicates that the distilled knowledge helps a lot though the architecture also achieve performance gains relative to the BiLSTM model with the contribution of more parameters.

\subsection{Application to DS-QA}

We then evaluate our method on open-domain question answering, which also requires understanding the long passage and predict the answer for a given question.
We use the OpenQA model\footnote{https://github.com/thunlp/OpenQA} \citep{lin2018denoising} as our baseline model, which applies a selector to filter passages, then a precise reader extracts the potential answers, finally aggregates these results to predict the final answer. We evaluate our method on two high-quality datasets, \emph{TriviaQA}\footnote{http://nlp.cs.washington.edu/triviaqa} (open-domain setting) \citep{joshi2017triviaqa} and \emph{SearchQA}\footnote{https://github.com/nyu-dl/SearchQA} \citep{dunn2017searchqa} with two metrics including ExactMatch (EM) and F1 scores.

\textbf{Implementation details:} For fairness, we use the pre-processed data\footnote{https://thunlp.oss-cn-qingdao.aliyuncs.com/OpenQA$\_$data.tar.gz} provided by \cite{lin2018denoising}.
We keep the same setup of hyper-parameters and training settings as that in OpenQA while some important details are as follows. We take the form of integration in equation (\ref{integrate_type_1}) to combine the passage selector with the gist detector and the $\lambda$ is set as 0.5. We feed the $\mathbf{v}_c$ through a linear function followed by multiplication with the question vector to produce the score for filtering passages and add it to original score produced by the OpenQA selector to predict the final passage score. For the reader, we directly add the predicted score of answer span with $\tilde{\mathbf{d}}$ in equation (\ref{gist_output}) with the form of combination formulated in equation (\ref{integrate_type_2}) to produce the final score, where the $\lambda^{'}$ is set as 0.2.
\begin{table}[tb]
	\centering
	\caption{EM and F1 scores on the test set of TriviaQa (open-domain setting) and SearchQA. In the top part, the results are taken from corresponding papers. In the medium part, we show the results of ablation experiments. OpenQA + ALL in the bottom denotes that the reader and selector are both combined with our gist detector.}
	\begin{tabular}{l c c c c c}
		\hline
		\multirow{2}*{\textbf{QA Models}} 
			& \multicolumn{2}{c}{\textbf{TriviaQA}} 
			&  \multicolumn{2}{c}{\textbf{SearchQA}} \\ \cline{2-5} 
		& \textbf{EM} & \textbf{F1}    & \textbf{EM}     & \textbf{F1}      \\ \hline
		BiDAF \citep{seo2016bidirectional}	&	- &	- &	28.6 &	34.6		\\
		AQA \citep{buck2018ask}			&	- &	- & 40.5	 &	47.4		\\
		R$^{3}$ \citep{wang2018r3}	 & 47.3 & 53.7 & 49.0 & 55.3 \\
		Re-Ranker \citep{wang2018evidence}	&	\textbf{50.6} &	57.3 & 57.0	 &	63.2	\\
		TraCRNet \citep{dehghani2019learning} & - & - & 52.9	 &	\textbf{65.1} \\
		OpenQA \citep{lin2018denoising} & 48.7 & 56.3  & 58.8  & 64.5 \\ 
		\hline
		OpenQA + selector with GD  &  49.4 & 57.1 & 59.2 &  64.8     \\
		OpenQA + reader with GD    &  49.2 & 56.5 & 59.0 &  64.8     \\
		\hline
		OpenQA + ALL &  50.3 & \textbf{57.6} & \textbf{59.5} &  \textbf{65.1}   \\
		\hline
	\end{tabular}
	\label{table_qa_performance}  
\end{table}
\begin{table}[tb]
	\centering
	\caption{Performance of passage selection on TriviaQA and SearchQA development set.}
	\begin{tabular}{l c c c c c c c}
		\hline
		\multirow{2}*{\textbf{QA Models}} 
			& \multicolumn{3}{c}{\textbf{TriviaQA}} 
			&  \multicolumn{3}{c}{\textbf{SearchQA}} \\ \cline{2-7} 
		& \textbf{Hit@1} & \textbf{Hit@3}    & \textbf{Hit@5} 
		& \textbf{Hit@1} & \textbf{Hit@3}    & \textbf{Hit@5}      \\ \hline
		OpenQA		&	43.4 &	51.5 &	54.5 &	59.1 &	68.7 &	76.3	\\
		OpenQA + GD	&	49.1 &	57.7 &	63.1 &	65.3 &	73.4 &	79.6	\\
		\hline
	\end{tabular}
	\label{table_qa_hit}  
\end{table}

\textbf{Results:} The results are shown in Table \ref{table_qa_performance}. The baseline model on two datasets performs much better with the help of our gist detector. We further conduct ablation experiments to evaluate the individual contribution when integrating the gist detector into selector and reader respectively. 
From the ablation experiments, we observe that the gist detector contributes to performance improvement though it is a bit modest while combined with the reader. But when we add both of them, the EM score on \emph{TriviaQA} rise about 1.6 which indicates that the transferred knowledge about finding out the salient information leads to better understanding of the long passages when the OpenQA model aggregates these information and predicts the final answer. In the Table \ref{table_qa_hit}, we show the performance in selecting related passages. We use Hit@N to represent the proportion of related passages being ranked in top-N. From the table, we found that the selector filters passages much precisely when in conjunction with the gist detector, which indicates our QA system can aggregate information among fewer passages and predict answers faster.

\subsection{Application to text style transfer}

We further conduct experiments on the text style transfer by changing the underlying sentiment of the text. In such task, the models compress the main information of the text into a fixed-size vector while segregating from the style information. We select the Cross-aligned AE\footnote{https://github.com/shentianxiao/language-style-transfer} \citep{shen2017style} and Adversarially Regularized Autoencoder (ARAE)\footnote{https://github.com/jakezhaojb/ARAE} \citep{kim2017adversarially} as our baseline models.
Despite the great process achieved by prior works, transferring styles among longer text is under-explored. In such scenario, the content preservation becomes the vital challenge. Therefore, it can further verify the effectiveness of our method.

\textbf{Implementation details:} We follow the setup of \cite{shen2017style} but remain reviews whose length are among 70 and 150 rather than no exceed 15, and finally collect 350K, 280K non-parallel datasets from Amazon and Yelp reviews respectively. We keep the same setup of hyper-parameters and training settings as that in Cross-aligned AE and ARAE. We combine the content vector with our gist detector as formulated in equation (\ref{integrate_type_1}), and the $\lambda$ is set to be 0.5.
To evaluate the model, we use 4 automatic metrics: (i) Acc: the accuracy of successfully changing the style into target style measured by a pre-trained classifier. Following \cite{shen2017style}, we use the TextCNN model as the classifier that achieves accuracy of 94.2\% and 95.7\% on Amazon and Yelp respectively. (ii) Cosine: we follow the setup of \cite{fu2018style} to measure the content preservation with cosine similarity. For fairness, we use the pre-trained 100d Glove vectors. (iii) Entity: we use the proportion of noun entity to measure the content consistent among the source and the generated text. (iv) PPL: the fluency of the generated text measured by the pre-trained language model on corresponding datasets.

\begin{table}[tb]
	\centering
	\caption{Automatic evaluation results on Amazon and Yelp datasets.}
	\begin{tabular}{c c c c c c c c c}
		\hline
		\multirow{2}*{\textbf{Models}} 
			& \multicolumn{4}{c}{\textbf{Amazon}} 
			&  \multicolumn{4}{c}{\textbf{Yelp}} \\ \cline{2-9} 
		& \textbf{Acc} & \textbf{Cosine} & \textbf{Entity} & \textbf{PPL} 
		& \textbf{Acc} & \textbf{Cosine} & \textbf{Entity} & \textbf{PPL}	\\ \hline
		Cross-aligned AE	 &	84.7\% &	0.46 &	26.13 &	34.67 &	89.5\% &	0.53 &	26.63 &	28.46	\\
		ARAE			 &	86.2\% &	0.57 &	31.37 &	36.36 &	89.3\% &	0.61 &	32.46 &	29.18	\\
		ARAE + GD		 &	91.0\% &	0.71 &	47.56 &	24.15 &	93.4\% &	0.73 &	49.04 &	21.43	\\
		\hline
	\end{tabular}
	\label{table_tst_performance}  
\end{table}

\begin{table}[tb]
	\centering
	\caption{Human evaluation on accuracy, content correlation and naturalness of the generated text.}
	\begin{tabular}{c c c c c c c}
		\hline
		\multirow{2}*{\textbf{Models}} 
			& \multicolumn{3}{c}{\textbf{Amazon}} 
			&  \multicolumn{3}{c}{\textbf{Yelp}} \\ \cline{2-7} 
		& \textbf{Acc} & \textbf{Correlation} & \textbf{Fluency} 
		& \textbf{Acc} & \textbf{Correlation} & \textbf{Fluency}	\\ \hline
		Cross-aligned AE	 & 56.4\% & 2.4 & 3.0 & 58.2\% & 2.7 & 3.1	 \\
		ARAE			 & 73.6\% & 2.8 & 3.3 & 74.1\% & 3.1 & 3.5	 \\
		ARAE + GD		 & 78.2\% & 3.7 & 3.5 & 78.6\% & 3.9 & 3.8 \\
		\hline
	\end{tabular}
	\label{table_tst_human_evaluation}  
\end{table}

\textbf{Results:} Table \ref{table_tst_performance} shows the quantitative evaluation on two datasets. From the results of Cross-aligned AE, we observe that it is easy for a model to get a high transfer accuracy but less content similarity by generating more target style words. Compared to the original ARAE model, the cosine similarity rises at least 0.12 and the proportion of entity preservation rises at least 16 while the accuracy also rises, which indicates that the distilled knowledge helps the ARAE model compress more important information into the context vector. Meanwhile, with the help of the gist detector, the augmented model generates much more fluent texts with smaller PPL value.
We conduct human evaluation to further evaluate the quality of the style transfer model.
The results are shown in Table \ref{table_tst_human_evaluation}.
We randomly select 1000 examples (500/500 positive/negative), and employ people to judge whether the text is converted to target style, evaluate the content correlation (0-5, 5 for most correlative) and the fluency (0-5, 5 for most natural). 
Further examples generated by all these models are provided in the supplementary material.

\section{Conclusion}

In this paper, we distill the knowledge about detecting salient information form the abstractive model and explore the way to transfer the distilled knowledge to other NLP tasks. 
The experimental results show that the simple implementation of the gist detector used in this paper significantly improves the performance of the baseline models for different NLP problems. Our approach is supposed to be further extended to many other problems in the field of NLP and the future works involve finding better strategies to integrate our gist detector into other models.

\bibliography{neurips_2020.bib}
\end{document}